\documentclass[letterpaper]{article} 
\usepackage{aaai2026}  
\usepackage{times}  
\usepackage{helvet}  
\usepackage{courier}  
\usepackage[hyphens]{url}  
\usepackage{graphicx} 
\usepackage{natbib}  
\usepackage{caption} 
\usepackage{algorithm}
\usepackage{algorithmic}

\usepackage{newfloat}
\usepackage{listings}
\DeclareCaptionStyle{ruled}{labelfont=normalfont,labelsep=colon,strut=off} 
\floatstyle{ruled}
\newfloat{listing}{tb}{lst}{}
\floatname{listing}{Listing}
\title{EcoAgent: An Efficient Device-Cloud Collaborative Multi-Agent Framework for Mobile Automation}
\author{
    Biao Yi\textsuperscript{\rm 1},
    Xavier Hu\textsuperscript{\rm 1},
    Yurun Chen\textsuperscript{\rm 1},
    Shengyu Zhang\textsuperscript{\rm 1}\thanks{Corresponding Author.},
    Hongxia Yang\textsuperscript{\rm 2},
    Fan Wu\textsuperscript{\rm 3}
}
\affiliations{
    \textsuperscript{\rm 1}Zhejiang University\\
    \textsuperscript{\rm 2}The Hong Kong Polytechnic University\\
    \textsuperscript{\rm 3}Shanghai Jiao Tong University\\

    \{yi\_biao, sy\_zhang\}@zju.edu.cn
}

\usepackage{bibentry}
\usepackage{booktabs}
\usepackage{multirow}
\usepackage{amsmath}
\usepackage{amssymb}
\usepackage{times}
\usepackage{helvet}
\usepackage{courier}
\usepackage{xcolor}

\begin{document}

\maketitle

\begin{abstract}
    To tackle increasingly complex tasks, recent research on mobile agents has shifted towards multi-agent collaboration. Current mobile multi-agent systems are primarily deployed in the cloud, leading to high latency and operational costs. A straightforward idea is to deploy a device–cloud collaborative multi-agent system, which is nontrivial, as directly extending existing systems introduces new challenges: (1) reliance on cloud-side verification requires uploading mobile screenshots, compromising user privacy; and (2) open-loop cooperation lacking device-to-cloud feedback, underutilizing device resources and increasing latency. To overcome these limitations, we propose \textbf{EcoAgent}, a closed-loop d\textbf{e}vice-\textbf{c}loud c\textbf{o}llaborative multi-agent framework designed for privacy-aware, efficient, and responsive mobile automation. EcoAgent integrates a novel reasoning approach, Dual-ReACT, into the cloud-based Planning Agent, fully exploiting cloud reasoning to compensate for limited on-device capacity, thereby enabling device-side verification and lightweight feedback. Furthermore, the device-based Observation Agent leverages a Pre-understanding Module to summarize screen content into concise textual descriptions, significantly reducing token usage and device-cloud communication overhead while preserving privacy. Experiments on AndroidWorld demonstrate that EcoAgent matches the task success rates of fully cloud-based agents, while reducing resource consumption and response latency.
\end{abstract}

\begin{links}
    \link{Code}{https://github.com/Yi-Biao/EcoAgent}
\end{links}

\section{Introduction}

\begin{figure*}[h]
  \centering
  \includegraphics[width=\linewidth]{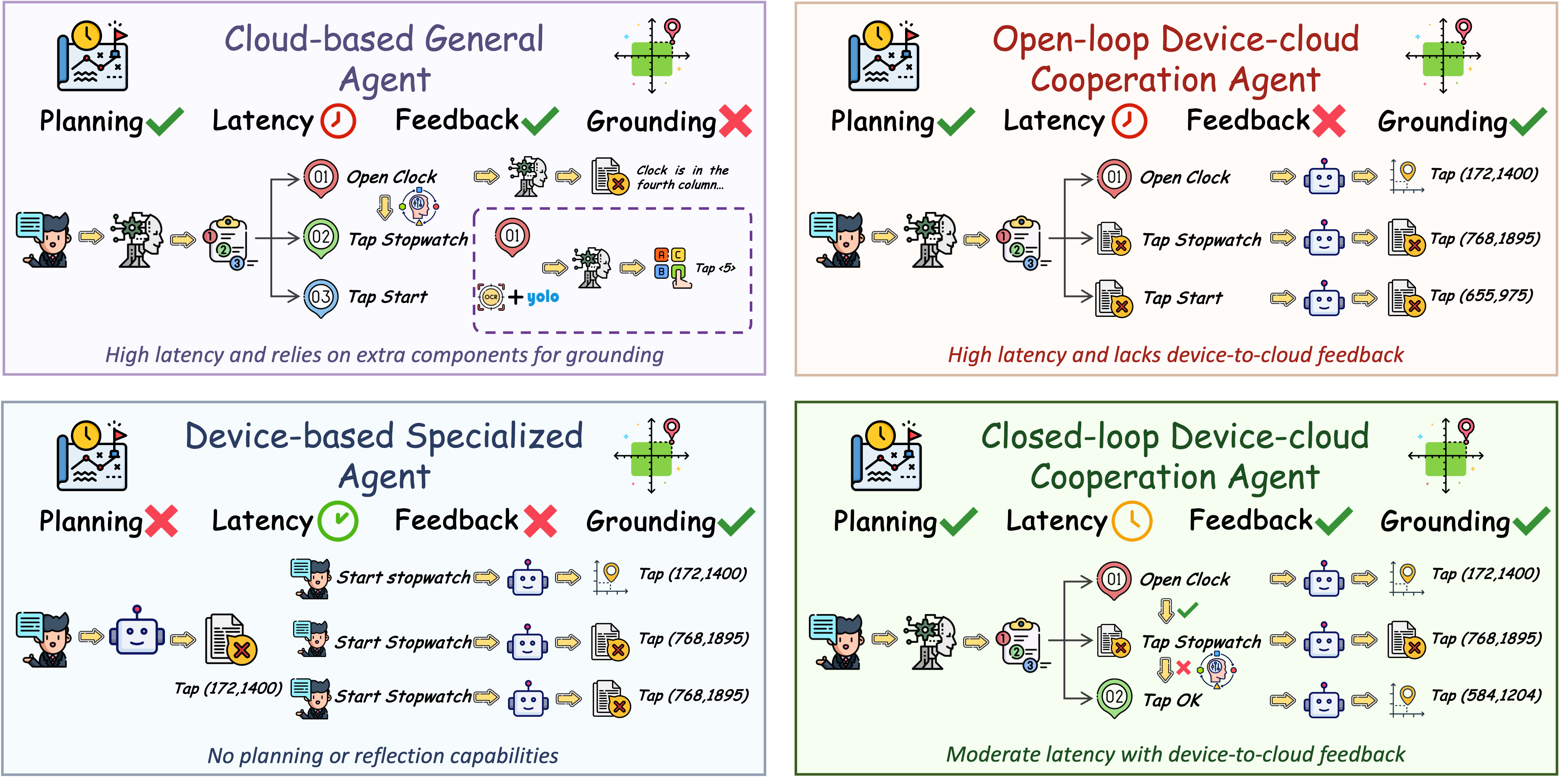}
  \caption{Comparison of four mobile agent architectures. Cloud-based General Agents exhibit strong planning abilities but suffer from high latency and rely on extra components for visual grounding. Device-based Specialized Agents achieve low latency and strong grounding but cannot plan or reflect. Open-loop Device-cloud Cooperation Agents enable grounding with precise control but lack device-to-cloud feedback. Closed-loop Device-cloud Cooperation Agents (\textbf{EcoAgent}) strike a balance by combining planning, feedback, and grounding, achieving moderate latency and improved adaptability.}
  \label{fig:compare}
\end{figure*}

With the rapid advancement of (Multimodal) Large Language Models ((M)LLMs), mobile agents capable of autonomously interacting with smartphones have garnered increasing attention \cite{hu2024agents,gao2024generalist,liu2025llm,wu2024foundations,zhang2024large,wang2024mobileagentbenchefficientuserfriendlybenchmark,chen2025aeiamnevaluatingrobustnessmultimodal,li2024effectsdatascaleui}.
A surge of mobile agents has been proposed based on (M)LLMs \cite{wang2024mobileagentautonomousmultimodalmobile,wu2024guiactionnarratordid}.
In recent years, research on mobile agents \cite{wang2024mobileagentv2mobiledeviceoperation,zhu2025mobamultifacetedmemoryenhancedadaptive,wang2025mobileagenteselfevolvingmobileassistant} has further leveraged the reasoning capabilities of (M)LLMs by delegating planning, execution, and reflection tasks to distinct agents. This collaborative approach has significantly enhanced the ability of mobile agents to tackle complex tasks.
However, due to the enormous parameter size of (M)LLMs, these agents predominantly operate in the cloud, resulting in high latency, high operational costs, and concerns about user privacy. 
Moreover, although some cloud-based agents \cite{qin2025ui,seed2025seed1_5vl} have been fine-tuned specifically for GUI tasks, the majority still rely on general-purpose (M)LLMs due to cost constraints. 
These cloud-based general agents cannot directly interact with UI elements and must depend on OCR and object detection modules for grounding, which further diminishes overall efficiency.


A natural solution is to deploy a device–cloud collaborative multi-agent system.
Many recent works \cite{wu2024osatlasfoundationactionmodel,cheng2024seeclickharnessingguigrounding,gou2025navigatingdigitalworldhumans} have made preliminary attempts in this direction by employing cloud-based agents for planning and device-based agents for grounding, yielding improved performance. 
However, they face two major challenges:
(1) Frequent uploading of mobile screenshots to the cloud for verification increases response latency and poses serious privacy risks.
(2) Such collaboration is typically open-loop, characterized by one-way cloud-to-device instructions without device-to-cloud feedback, failing to fully exploit the potential of device–cloud cooperation to reduce operational costs and latency.

To address these challenges, we introduce \textbf{EcoAgent}, a closed-loop device–cloud collaborative multi-agent framework for privacy-aware, efficient, and responsive mobile automation. EcoAgent consists of a cloud-based \textbf{Planning Agent}, a device-based \textbf{Execution Agent}, and a device-based \textbf{Observation Agent}, working together in a feedback loop. As illustrated in Figure~\ref{fig:compare}, EcoAgent achieves moderate latency and improved adaptability compared to existing agent architectures.

To support device-to-cloud feedback, the Planning Agent employs Dual-ReACT, a two-stage reasoning framework that first performs global ReACT-based planning over the instruction and initial screen, followed by local ReACT-based generation of step-by-step actions and corresponding expectations.
On the device, the Execution Agent performs each step in the plan, while the lightweight Observation Agent compares the current screen with the expectation to verify execution status—without requiring strong reasoning. This enables efficient on-device verification and device-to-cloud feedback, with cloud assistance needed only when execution fails.
To further reduce latency and protect user privacy, the Observation Agent incorporates a Pre-Understanding Module that compresses screenshots into concise textual summaries before transmission, significantly reducing token consumption and communication overhead.
Finally, the Planning Agent integrates Memory and Reflection modules to support replanning in response to device-side feedback, enabling the system to adaptively recover from failure. 

We evaluate EcoAgent on the widely-used benchmark AndroidWorld. Experimental results demonstrate that EcoAgent not only effectively handles complex mobile tasks, but also significantly reduces operational costs and response latency. 
Our contributions are summarized as follows:
\begin{itemize}
    \item We propose \textbf{EcoAgent}, a closed-loop device–cloud collaborative framework that balances cloud-based reasoning with lightweight on-device verification, enabling privacy-aware and efficient mobile automation.
    \item We introduce \textbf{Dual-ReACT}, a novel two-stage planning mechanism that combines global and local ReACT reasoning to generate executable action plans with explicit expectations.
    \item We design a \textbf{Pre-Understanding Module} within the Observation Agent to compress visual inputs into textual summaries, protecting user privacy and significantly reducing device–cloud communication overhead.
\end{itemize}

\section{Related Work}

\begin{figure*}[h]
  \centering
  \includegraphics[width=\linewidth]{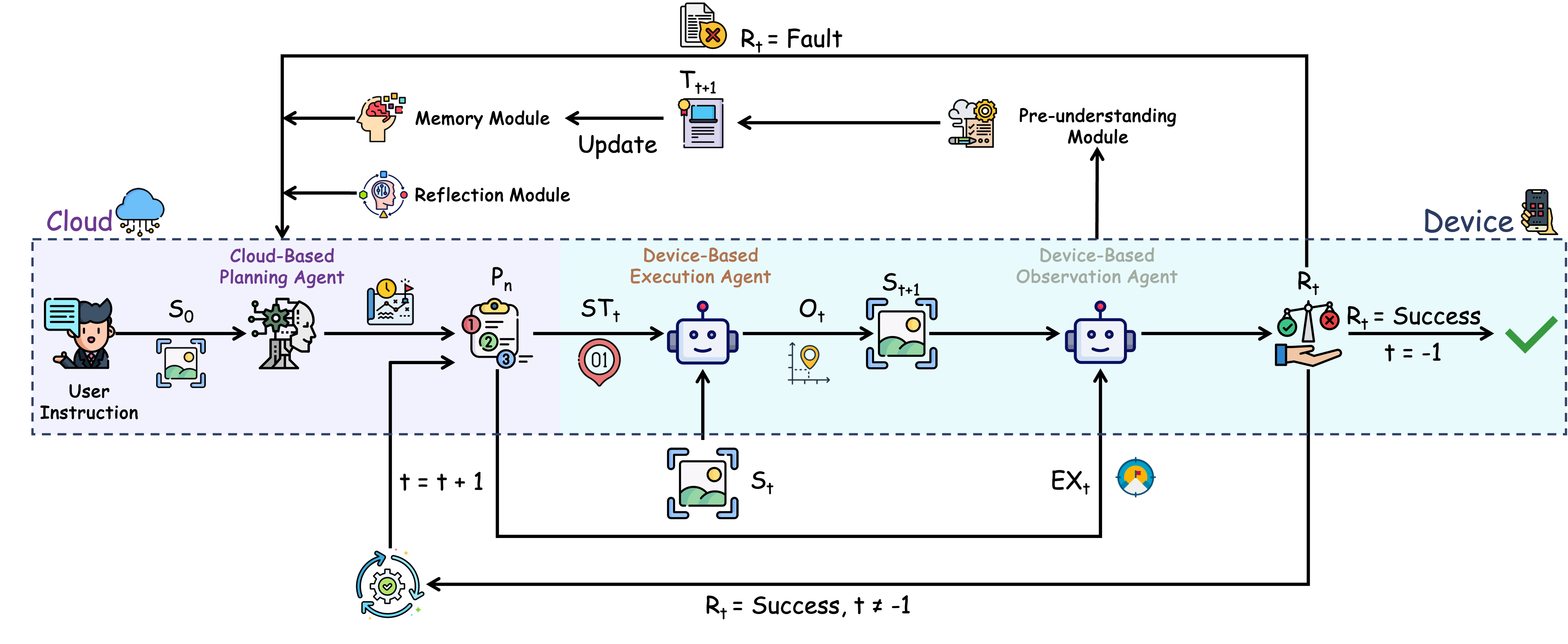}
  \caption{An overview of the EcoAgent framework. The Cloud-Based Planning Agent generates a task plan, which is executed and verified step-by-step by on-device agents. Feedback and pre-understood screen text from the Observation Agent enable dynamic replanning, forming a closed-loop system for efficient and adaptive task execution.}
  \label{fig:overview}
\end{figure*}

\subsection{Cloud-Based Mobile Agent}

The rapid advancement of (M)LLMs has empowered mobile agents with strong capabilities in perception, reasoning, and decision-making. However, due to their massive model size and computational demands, (M)LLM-based mobile agents are typically deployed in the cloud, leading to significant latency and high operational costs, as illustrated in Figure~\ref{fig:compare}.
Moreover, as the vast majority of MLLMs are trained primarily on general natural images, cloud-based general agents \cite{zhang2023appagentmultimodalagentssmartphone,liu2024groundingdinomarryingdino,wang2023enabling,wen2024droidbotgptgptpowereduiautomation,taeb2024axnav,lee2024exploreselectderiverecall,Huang_2025,zhang2024mobileexpertsdynamictoolenabledagent,liu2024seeingbelievingvisiondrivennoncrash,li2024appagentv2advancedagent,jiang2025appagentxevolvingguiagents,yan2023gpt4vwonderlandlargemultimodal,Song_2024} struggle to accurately perceive and interact with mobile screens, often requiring extra components for UI element recognition and localization. 
For instance, Mobile-Agent \cite{liu2024groundingdinomarryingdino} integrates DINO \cite{liu2024groundingdinomarryingdino} and CLIP \cite{radford2021learning} recognition components, along with OCR-extracted text, to generate textual descriptions of screens, aiding MLLMs in screen understanding.

\subsection{Device-Based Mobile Agent}

To address the limitations of cloud-based general agents, recent studies \cite{hong2024cogagentvisuallanguagemodel,lin2024showuivisionlanguageactionmodelgui,liu2025infiguiagentmultimodalgeneralistgui,wu2024osatlasfoundationactionmodel,wen2024autodroidv2boostingslmbasedgui,dai2025advancingmobileguiagents} have focused on fine-tuning multimodal Small Language Models ((M)SLMs) to enable fast and accurate mobile interactions. For example, ShowUI \cite{lin2024showuivisionlanguageactionmodelgui} proposes a vision-language-action model that integrates UI-guided visual token selection and interleaved vision-language-action streaming for efficient GUI manipulation. In addition, recent work \cite{Wang_2024} has demonstrated the feasibility of deploying 7B MSLM directly on mobile devices, showing their potential for on-device automation. However, as illustrated in Figure~\ref{fig:compare}, due to their limited model capacity, these device-based specialized agents often struggle with effective planning and reflection, making it difficult to handle complex mobile tasks.

\subsection{Multi-Agent Mobile Agent}


Due to the challenges single agents face in handling complex tasks, recent studies \cite{wang2024mobileagentv2mobiledeviceoperation,zhu2025mobamultifacetedmemoryenhancedadaptive,wang2025mobileagenteselfevolvingmobileassistant,chen2025harmonyguardsafetyutilityweb,chen2025graph2evalautomaticmultimodaltask,wang2025efficientagentsbuildingeffective} have explored enhancing agent capabilities through multi-agent collaboration. For example, M3A \cite{rawles2024androidworld} employs a dual-agent design, where one agent is responsible for decision-making and execution, while the other verifies execution status and performs reflection. However, cloud-based multi-agent systems like M3A further exacerbate latency and operational cost issues. A natural solution is device–cloud collaborative multi-agent systems. For instance, UGround \cite{gou2025navigatingdigitalworldhumans} extends M3A by introducing an device-based grounding agent for execution, which improves task success rates. However, verification and planning still occur in the cloud, requiring frequent screenshot uploads. As illustrated in Figure~\ref{fig:compare}, such open-loop collaborations continue to suffer from high latency, high costs, and privacy concerns due to one-way cloud-to-device instruction flow without device-to-cloud feedback.

\section{The EcoAgent Framework}

\begin{figure*}[h]
  \centering
  \includegraphics[width=\linewidth]{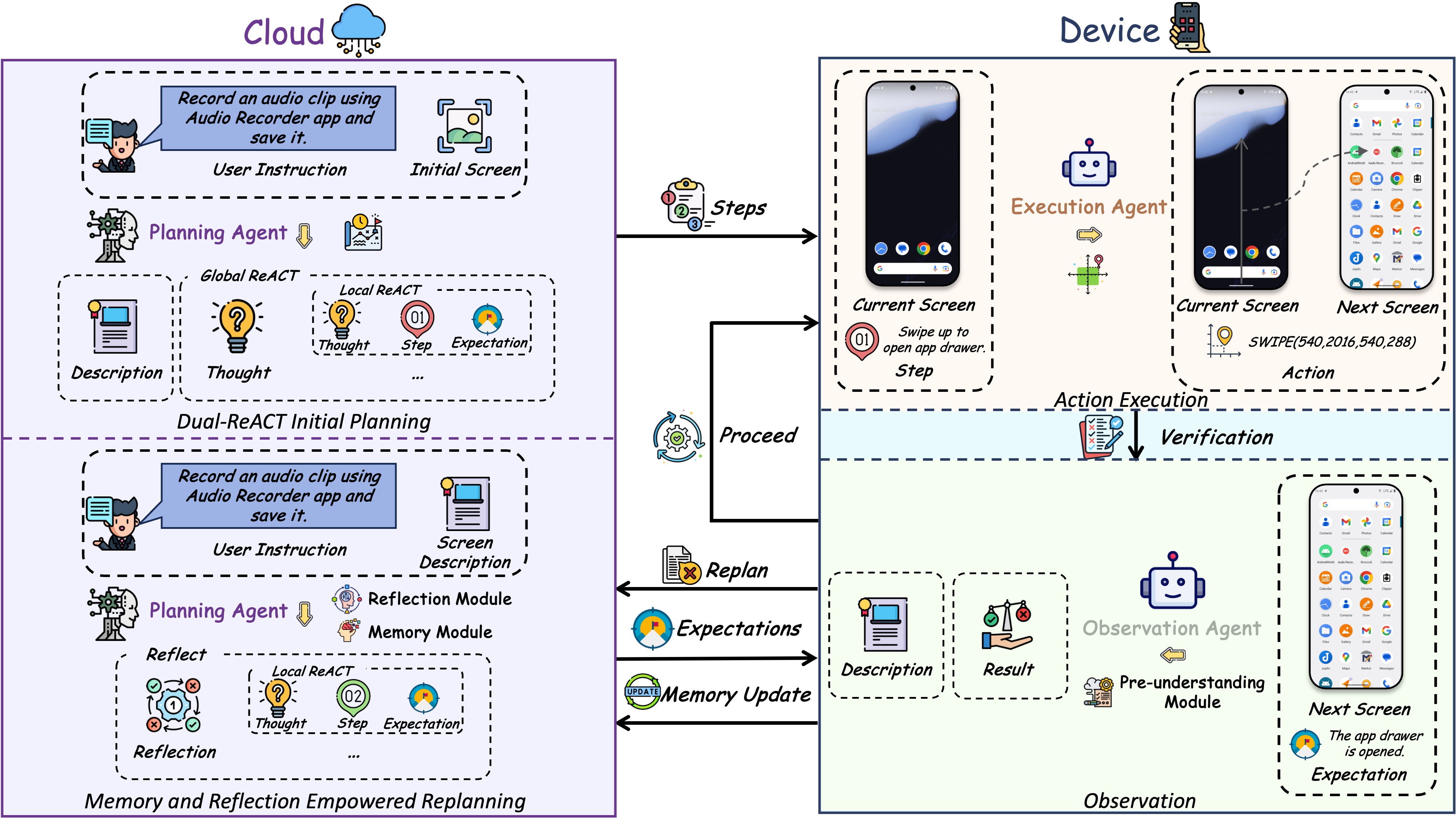}
  \caption{Illustration of the EcoAgent workflow. The cloud-based Planning Agent is responsible for Dual-ReACT initial planning as well as Memory and Reflection empowered replanning. The device-based Execution Agent and Observation Agent handle action execution and verification, respectively. Together, these three agents collaborate to realize a closed-loop workflow.}
  \label{fig:workflow}
\end{figure*}

\begin{algorithm}[h!]
\caption{Workflow of the \textbf{EcoAgent}}
\label{algorithm:Workflow of EcoAgent}
\textbf{Input}: User instruction $Ins$, Initial screen state $S_0$\\
\textbf{Output}: Final system state $S_f$

\begin{algorithmic}[1]
\STATE \textbf{Step 1: Initial Planning.} 
\STATE $P_0 \leftarrow DualReACT(Ins, S_0)$ 
\STATE \textbf{Step 2: Execution and Observation.}
\FOR{each $(ST_t, EX_t)$ in $P_0$}
\STATE Execute current step: 
\STATE $O_t \leftarrow EA(S_t, ST_t)$
\STATE $S_{t+1} \leftarrow O_t(S_t)$
\STATE Verify result: 
\STATE $R_t \leftarrow OA(S_{t+1}, EX_t)$
\STATE Compress screen: 
\STATE $T_{t+1} \leftarrow PreUnderstand(S_{t+1})$
\STATE Update memory: 
\STATE $Memory \leftarrow Memory \cup \{T_{t+1}\}$
\IF{$R_t == \texttt{Fail}$}
\STATE \textbf{Step 3: Reflection and Replanning.}
\STATE Replan: 
\STATE $P_n \leftarrow Reflection(Ins, P_{n-1}, Memory)$
\STATE Restart execution with updated plan $P_n$
\STATE \textbf{break}
\ENDIF
\ENDFOR
\STATE Final system state: $S_f \leftarrow S_t$
\STATE \textbf{Return} $S_f$
\end{algorithmic}
\end{algorithm}

In this section, we provide a detailed overview of the EcoAgent framework. EcoAgent adaptively coordinates cloud-based and device-based agents to execute user instructions, as illustrated in Figure~\ref{fig:overview}. The framework consists of three specialized agents: a cloud-based Planning Agent and two device-based agents, the Execution Agent and the Observation Agent.

The overall workflow of EcoAgent is illustrated in Algorithm 1 and proceeds as follows. Firstly, the Planning Agent performs Dual-ReACT initial planning based on user instruction $Ins$ and initial screen $S_0$, generating an initial plan $P_0$ consisting of a sequence of steps $ST_t$ and their corresponding expectations $EX_t$. The Execution Agent on the device side executes actions $O_t$ based on the given step $ST_t$ and the current screen $S_t$, obtains the next screen $S_{t+1}$. After each action, the Observation Agent evaluates whether the step was successfully executed by comparing the next screen $S_{t+1}$ against the corresponding expectation $EX_t$ to obtain the result $R_t$. To reduce token consumption and device-cloud communication overhead, the Observation Agent incorporates a Pre-Understanding Module, which converts high-token-cost images $S_{t+1}$ into compact textual representations $T_{t+1}$ and updates them to the memory module. If a step fails, the Planning Agent conducts Memory and Reflection empowered replanning, learning from the failure of the previous plan $P_{n-1}$ to generate a new one $P_{n}$, and continues this process until the task is successfully completed.

\subsection{Cloud-Based Planning Agent}

Despite the difficulty MLLMs face in accurately identifying and grounding individual elements within mobile screens, their strong long-horizon planning and reflection capabilities make them well-suited for understanding screen context and adapting plans to dynamic environments. These strengths allow MLLMs to effectively decompose and solve complex mobile tasks—capabilities that MSLMs often lack. To leverage these strengths, we design a Cloud-based Planning Agent powered by MLLMs, which transforms high-level user instructions into adaptive and verifiable step-by-step plans. This agent operates through two primary mechanisms: Dual-ReACT Initial Planning and Memory and Reflection Empowered Replanning.

\textbf{Dual-ReACT Initial Planning.}  
Open-loop device–cloud cooperation agents often relies on cloud-side verification due to limited on-device capabilities, which makes it difficult to directly analyze execution outcomes from screenshots. However, general-purpose multimodal MSLMs are adept at describing screen content and judging whether it aligns with a given description. This motivates transforming complex task analysis into a simpler judgment by fully leveraging powerful cloud-side reasoning.

To this end, we propose a novel reasoning paradigm, Dual-ReACT, which extends the original ReACT framework \cite{yao2023reactsynergizingreasoningacting} by applying reasoning and acting at both global and local levels.

As illustrated in Figure~\ref{fig:workflow}, given a user instruction $Ins$ and an initial screen state $S_0$, the Planning Agent first generates a global screen description to capture the overall context. Based on this, it conducts a Global ReACT process to derive a high-level execution plan by decomposing the task into intermediate subgoals. For each subgoal, a Local ReACT reasoning step determines how to achieve it, generating a concrete action step $ST_t$ along with its expected outcome $EX_t$.

The resulting plan can be formulated as:
\begin{align*}
       P_0 &= \text{GlReACT}(Ins, S_0) \\
       &= \left\{ \text{LoReACT}(ST_1, EX_1), \ldots, \text{LoReACT}(ST_t, EX_t) \right\} 
\end{align*}

This plan is then transmitted to the device, enabling effective device-to-cloud feedback and realizing a closed-loop.

\textbf{Memory and Reflection Empowered Replanning.}
When execution fails, the Planning Agent initiates a memory-driven reflection process as shown in Figure~\ref{fig:workflow}. The Memory Module stores screen and action trajectories, which are then used as contextual input to the Reflection Module, enabling the system to analyze error trajectories and adaptively revise the task plan:
\begin{displaymath}
P_n = \text{Reflection}(Ins, P_{n-1}, \text{Memory})
\end{displaymath}

This mechanism is inspired by prior work on LLM-based reflection for decision making~\cite{shinn2023reflexion}, which demonstrates the benefits of leveraging past experiences to iteratively refine agent behavior.

\subsection{Device-Based Execution Agent}

The Execution Agent, deployed on the device, is responsible for executing operation $O_t$ based on the current screen state $S_t$ and the corresponding step $ST_t$ generated by the Planning Agent, which is denoted as:

\begin{displaymath}
O_t = EA(S_t, ST_t)
\end{displaymath}
where $EA$ represents the fine-tuned MSLM of the Execution Agent.

Leveraging fine-tuned MSLMs with strong grounding capabilities, the Execution Agent achieves high accuracy and low latency in input operations, which is often lacking in MLLM-based agents.

In addition to basic input execution, we design navigation operations that fully leverage the long-horizon planning and reflection capabilities of the cloud-based Planning Agent, allowing EcoAgent to promptly exit failure states and re-execute the correct actions. Below is a detailed description of the operation space:

\begin{itemize}
    \item \textbf{Input Operations}:
    \begin{itemize}
        \item \texttt{Tap(x, y)}: Tap the screen at coordinate $(x, y)$.
        \item \texttt{Swipe(x\textsubscript{start}, y\textsubscript{start}, x\textsubscript{end}, y\textsubscript{end})}: Swipe from $(x\textsubscript{start}, y\textsubscript{start})$ to $(x\textsubscript{end}, y\textsubscript{end})$.
        \item \texttt{LongPress(x, y)}: Long press at coordinate $(x, y)$.
        \item \texttt{InputText(text)}: Input the string \texttt{text} into the currently focused field.
        \item \texttt{DeleteText()}: Delete all content in the input field.
        \item \texttt{OpenApp(app name)}: Open specific application.
    \end{itemize}
    
    \item \textbf{Navigation Operations}:
    \begin{itemize}
        \item \texttt{PressBack()}: Simulate the system back button.
        \item \texttt{PressHome()}: Simulate the system home button.
    \end{itemize}
\end{itemize}

To improve the flexibility of input operations, we extend the action space with a new deletion action. We observe that existing mobile agents lack an explicit mechanism for text deletion, making it difficult to re-enter text once incorrectly input. To address this issue, we introduce the \texttt{DeleteText()} operation, which clears all content in the current input field located. This enhancement allows EcoAgent to effectively perform rename tasks. 

\subsection{Device-Based Observation Agent}

Although the device-based Execution Agent can effectively recognize and locate screen elements, we find that due to the small parameter size of the fine-tuned MSLMs and the large amount of GUI data used for fine-tuning, their textual capabilities are significantly degraded, leading to challenges in assessing the success of each action. While cloud-based MLLMs could address this issue, they introduce significant latency and cost. To overcome this, we design the device-based Observation Agent, which employs general MSLMs not fine-tuned on large-scale GUI data, enabling accurate understanding of screen semantics and instruction-aligned expectations. Specifically, as shown in Figure~\ref{fig:workflow} the Observation Agent is responsible for determining whether a step is successfully executed by comparing the post-execution screen state $S_{t+1}$ with the expected outcome $EX_t$, denoted as:

\begin{displaymath}
    R_t = OA(S_{t+1}, EX_t)
\end{displaymath}
where $R_t$ represents the result of execution verification, and $OA$ represents the general MSLM of the Observation Agent.

\textbf{Pre-Understanding Module.} Screen history plays a crucial role in helping the Planning Agent identify error causes and replan accordingly. However, directly storing raw screenshots in the memory module presents two key challenges: (1) the original screen content must be transmitted to the cloud, potentially leading to user privacy leakage; and (2) image transmission introduces high device–cloud latency and significantly increases token consumption when processed by MLLMs. Moreover, effective replanning does not require access to all screen details, but only the screen transition trajectory—that is, the high-level semantic changes across screens that reflect execution progress or failure.

To address these issues, we introduce a Pre-Understanding Module within the Observation Agent. This module transforms the raw screen state $S_{t+1}$ into a compact textual representation $T_{t+1}$:

\begin{displaymath}
T_{t+1} = \text{PreUnderstand}(S_{t+1})
\end{displaymath}

Through the Pre-Understanding Module, screen images that typically consume over 1400 tokens per image are encoded into concise textual descriptions requiring only 50–150 tokens. This significantly reduces device–cloud communication overhead and MLLM token consumption. Moreover, by converting full-screen content into simplified textual descriptions rather than transmitting raw images, the risk of user privacy leakage is substantially mitigated.

\begin{table*}[t]
\centering
\begin{tabular}{clllc}
\toprule
\textbf{Agent Architecture} &\textbf{Agent}&\textbf{Type}&\textbf{Foundation Model(s)}  & \textbf{SR (\%)} \\
\midrule
\multirow{4}{*}{Device-based} & ShowUI & Single & ShowUI (2B) & 7.0 \\
& InfiGUIAgent & Single & InfiGUIAgent (2B) & 9.0\\
& OS-Atlas & Single & OS-Atlas-Pro (4B) & 4.3\\
& V-Droid & Single & V-Droid (8B) &59.5\\
\midrule
\multirow{3}{*}{Cloud-based} &AppAgent & Single        &  GPT-4o     &         11.2        \\
&M3A    & Multi          &  GPT-4o * 2      &     28.4       \\
& Agent S2 & Multi & GPT-4o * 4 & 54.3 \\
\midrule
\multirow{2}{*}{Open-loop Device-Cloud}& UGround-V1-2B & Multi & GPT-4o * 2, UGround-V1-2B & 32.8\\
&UGround-V1-7B & Multi & GPT-4o * 2, UGround-V1-7B & 44.0\\
\midrule
\multirow{2}{*}{\textbf{Closed-loop Device-Cloud}}&EcoAgent (ShowUI)  & Multi       &  GPT-4o, ShowUI (2B), Qwen2-VL-2B      &    25.6          \\
&EcoAgent (OS-Atlas) & Multi       &  GPT-4o, OS-Atlas-Pro (4B), Qwen2-VL-2B     &  27.6            \\
\bottomrule
\end{tabular}
\caption{Success Rate Comparison of Mobile Agents on AndroidWorld.}
\label{tab:sr}
\end{table*}

\begin{table}[H]
\centering
\begin{tabular}{clcc}
\toprule
\textbf{Architecture}&\textbf{Agent} & \textbf{MC} & \textbf{MT} \\
\midrule
Device-based&ShowUI & 0 & 0\\
\midrule
\multirow{2}{*}{Cloud-based}&AppAgent        &          6.46        &      15309       \\
&M3A             &     13.39        &      87469       \\
\midrule
Open-loop & UGround-V1-2B &12.21 &45192\\
\midrule
\multirow{2}{*}{\textbf{Closed-loop}}&EcoAgent (ShowUI)         &    1.87         &   3545          \\
&EcoAgent (OS-Atlas)        &  1.53            &      3240        \\
\bottomrule
\end{tabular}
\caption{Operational Costs Comparison of Mobile Agents.}
\label{tab:OC}
\end{table}
\section{Experiments}

In this section, we compare EcoAgent with other mobile agents of different architectures in terms of performance, operational costs, and latency. 
In addition, we conduct an ablation study to validate the contribution of the proposed components within EcoAgent.

\subsection{Benchmark}

\textbf{AndroidWorld} \cite{rawles2024androidworld} is a dynamic benchmarking environment designed to evaluate autonomous agents on Android devices. It offers 116 programmatic tasks across 20 real-world Android applications. The benchmark runs on a live Android emulator, specifically configured with the Pixel 6 device model and Android 13 (API Level 33), ensuring a consistent and realistic testing environment. While other online benchmarks such as \textbf{MobileAgentBench} \cite{wang2024mobileagentbenchefficientuserfriendlybenchmark} and \textbf{AndroidLab} \cite{xu2024androidlabtrainingsystematicbenchmarking} also cover a wide range of real-world tasks, they rely heavily on human evaluation to determine success or failure. This introduces challenges in reproducibility, scalability, and evaluation latency. In contrast, AndroidWorld offers fully programmatic evaluation, enabling consistent, automated, and reproducible measurements of agent performance without requiring manual intervention.

\subsection{Baselines}

To ensure a fair and comprehensive evaluation of EcoAgent, we compare it against several mobile agents of different architectures, including the device-based agents ShowUI \cite{lin2024showuivisionlanguageactionmodelgui}, OS-Atlas \cite{wu2024osatlasfoundationactionmodel}, InfiGUIAgent \cite{liu2025infiguiagentmultimodalgeneralistgui} and V-Droid \cite{dai2025advancingmobileguiagents}; the cloud-based single-agent AppAgent \cite{zhang2023appagentmultimodalagentssmartphone} and MobileAgent \cite{wang2024mobileagentautonomousmultimodalmobile}; the cloud-based multi-agent M3A \cite{rawles2024androidworld} and Agent S2 \cite{agashe2025agents2compositionalgeneralistspecialist}; and the open-loop device–cloud collaborative multi-agent system UGround \cite{gou2025navigatingdigitalworldhumans}, CogAgent \cite{hong2024cogagentvisuallanguagemodel} and AutoDroid \cite{wen2024autodroid}.

For fairness, the cloud-based agents for all baselines are implemented using GPT-4o. For practical future deployment on mobile devices, EcoAgent adopts a lightweight design: the Observation Agent is built upon Qwen2-VL-2B, and the Execution Agent integrates the SOTA device-side models ShowUI (2B) and OS-Atlas-Pro (4B).

\subsection{Metrics}

\begin{table*}[t]
\centering
\begin{tabular}{ccccccccc}
\toprule
\textbf{Executor} & \textbf{Planner} & \textbf{Observer} &\multicolumn{3}{c}{\textbf{Ablation Setting}} & \multicolumn{3}{c}{\textbf{AndroidWorld}}  \\
\cmidrule(lr){4-6} \cmidrule(lr){7-9}
\textbf{Model}& \textbf{Model}& \textbf{Model}&\textbf{Executor} & \textbf{Planner} & \textbf{Observer} & \textbf{SR (\%)}  & \textbf{MC} & \textbf{MT} \\
\midrule
\multirow{3}{*}{ShowUI}&\multirow{3}{*}{GPT-4o}&\multirow{3}{*}{Qwen2-VL-2B}&\checkmark &   &  & 7.0  & 0 & 0 \\
&&&\checkmark & \checkmark &  & 15.5  & 1 & 2149 \\
&&&\checkmark & \checkmark & \checkmark & 25.6  & 1.87 & 3545\\
\midrule
\multirow{3}{*}{OS-Atlas}&\multirow{3}{*}{GPT-4o}&\multirow{3}{*}{Qwen2-VL-2B}&\checkmark &  &  & 4.3   & 0 & 0 \\
&&&\checkmark & \checkmark & & 19.0  & 1 & 2181 \\
&&&\checkmark & \checkmark & \checkmark & 27.6 &  1.53& 3240\\
\bottomrule
\end{tabular}
\caption{Ablation results on AndroidWorld with different combinations of Execution Agent, Planning Agent, and Observation Agent.}
\label{tab:ablation}
\end{table*}

We evaluate mobile agent performance along three dimensions: task success rate (SR), operational costs, and latency. The task success rate is defined as the percentage of tasks successfully completed by the agent. To assess operational costs, we consider two key metrics: Average MLLM Calls (MC) and Average MLLM Tokens (MT). MC denotes the average number of times the agent requests assistance from the cloud-side MLLM per task, while MT represents the average number of tokens consumed by the cloud model per task. Latency refers to the average time taken for the agent to complete each execution step.

\subsection{Evaluation}

\subsubsection{Evaluation on Task Success Rate} 

Table ~\ref{tab:sr} presents the success rate (SR) comparison of various mobile agents on the AndroidWorld benchmark. Among all baselines, V-Droid achieves the highest SR at 59.5\%, demonstrating its strong capability as a specialized device-based agent. Our proposed \textbf{EcoAgent} achieves SRs of 25.6\% and 27.6\% when using ShowUI and OS-Atlas as Execution Agents, respectively. This surpasses all single-agent baselines and even matches the performance of M3A (28.4\%), a cloud-based dual-agent system using two GPT-4o models. This demonstrates that EcoAgent’s closed-loop design, which leverages lightweight on-device agents and cloud reflection via Dual-ReACT, can achieve a similar level of task success with significantly lower cloud reliance. Importantly, EcoAgent is orthogonal to V-Droid, and future work can explore integrating V-Droid as the Execution Agent, potentially combining its strong low-level execution ability with EcoAgent’s reasoning, reflection, and planning capabilities.

\subsubsection{Evaluation on Operational Costs} 

Table~\ref{tab:OC} presents the operational costs comparison of different mobile agents on the AndroidWorld benchmark. Compared to the cloud-based single-agent baseline AppAgent, EcoAgent (OS-Atlas) reduces MC by 76\% and MT by 79\%. When compared to the cloud-based dual-agent M3A, EcoAgent achieves even more significant savings: MC is reduced by 89\% and MT by 96\%. Although UGround-V1-2B introduces an device-based grounding agent to reduce cloud communication, it still requires 6.5$\times$ more MC and 14$\times$ more MT than EcoAgent (OS-Atlas). These results demonstrate that EcoAgent’s closed-loop device–cloud collaboration substantially reduces operational costs, making it a more practical and scalable solution for real-world mobile automation.

\subsubsection{Evaluation on Latency}

Table~\ref{tab:LT} shows the latency comparison of mobile agents. Device-based agents, like ShowUI and V-Droid, have the lowest latencies at 1.2 seconds and 3.0 seconds, respectively. Cloud-based agents like AppAgent, MobileAgent, and M3A exhibit much higher latencies, ranging from 7.1 seconds to 15.9 seconds, due to frequent cloud interactions. Open-loop device-loud cooperation agents fall in between with latencies from 4.9 seconds to 18.2 seconds. EcoAgent (ShowUI) achieves a latency of 3.9 seconds, significantly lower than other cloud-based and open-loop device-loud cooperation agents, demonstrating the efficiency of its closed-loop device-cloud collaboration.

\begin{table}[h]
\centering
\begin{tabular}{clc}
\toprule
\textbf{Architecture}& \textbf{Agent} & \textbf{Latency (s)}\\
\midrule
\multirow{2}{*}{Device-based}&ShowUI & 1.2 \\
&V-Droid & 3.0 \\
\midrule
\multirow{3}{*}{Cloud-based}&APPAgent& 7.1 \\
&MobileAgent & 15.9 \\
&M3A & 15.3 \\
\midrule
\multirow{3}{*}{Open-loop}&UGroud-V1-2B & 18.2 \\
&CogAgent & 6.8 \\
&AutoDroid & 4.9 \\
\midrule
\textbf{Closed-loop}&EcoAgent (ShowUI) & 3.9 \\
\bottomrule
\end{tabular}
\caption{Latency Comparison of Mobile Agents.}
\label{tab:LT}
\end{table}

\subsection{Ablation Study}

Table~\ref{tab:ablation} reports the ablation results of EcoAgent on AndroidWorld under different agent combinations.

When only the Execution Agent is used, the success rate (SR) is very low (7.0\% for ShowUI and 4.3\% for OS-Atlas), demonstrating that relying solely on the Execution Agent leads to poor performance on complex tasks. 

Introducing the Planning Agent with the Dual-ReACT framework substantially improves performance, increasing SR to 15.5\% (ShowUI) and 19.0\% (OS-Atlas). This validates the effectiveness of Dual-ReACT in providing structured task decomposition and reasoning.

Further adding the Observation Agent enables device-to-cloud feedback and integrates the Pre-Understanding Module. This enhances the SR to 25.6\% (ShowUI) and 27.6\% (OS-Atlas), demonstrating the crucial role of the Observation Agent. Notably, the increase in MC and MT is moderate. MC rises from 1 to approximately 1.5 to 1.9, while MT increases from around 2,100 to between 3,200 and 3,500. This indicates that the Pre-Understanding Module effectively reduces communication and computation costs while enabling closed-loop collaboration.

\section{Conclusion}


In this paper, we propose EcoAgent, a closed-loop device-cloud collaborative multi-agent framework for privacy-aware, efficient mobile automation. Experiments on AndroidWorld show EcoAgent reduces cloud reliance and performs robustly on complex tasks. While full on-device deployment of MSLMs remains challenging, we expect EcoAgent to become increasingly deployable as edge AI hardware and model efficiency continue to advance.

\section{Acknowledgments}
This work was supported by National Science and Technology Major Project (2022ZD0119100), National Natural Science Foundation of China (No. 62402429, U24A20326, 62441236), Key Research and Development Program of Zhejiang Province (No. 2025C01026), Ningbo Yongjiang Talent Introduction Programme (2023A-397-G) and Young Elite Scientists Sponsorship Program by CAST (2024QNRC001). The author also gratefully acknowledges the support of the Zhejiang University Education Foundation Qizhen Scholar Foundation.

\bibliography{aaai2026}

\clearpage
\onecolumn
\section{Appendix}

\begin{figure*}[h!]
  \centering
  \includegraphics[width=\linewidth]{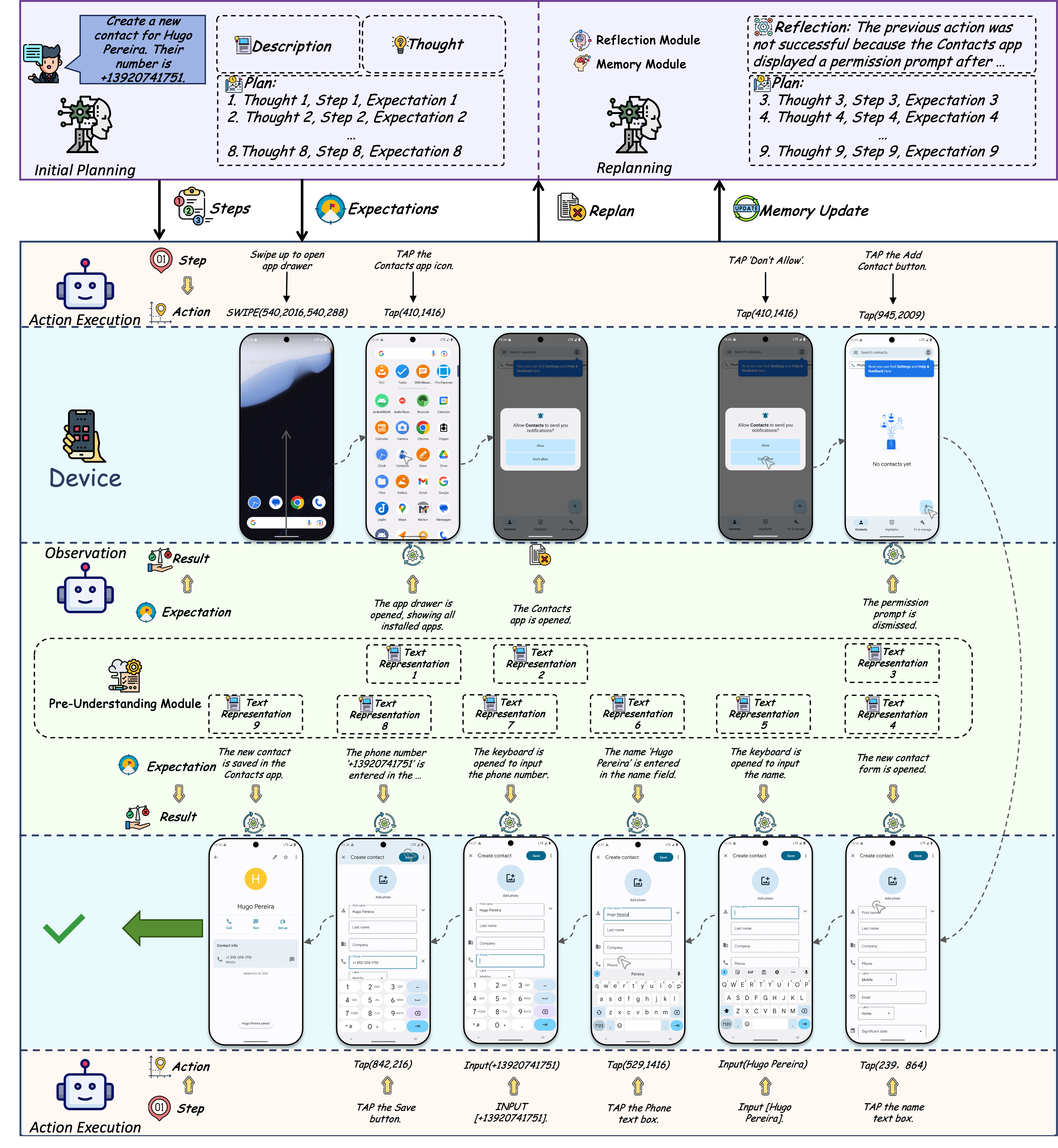}
  \caption{A complete case of EcoAgent executing the ContactsAddContact task in AndroidWorld. The figure illustrates the full interaction loop among Planning, Execution, and Observation Agents across multiple steps until task completion.}
  \label{fig:casestudy}
\end{figure*}

\subsection{Case Study}

Figure~\ref{fig:casestudy} illustrates the complete operational process of \textbf{EcoAgent} in executing complex multi-step tasks on mobile devices.  The Planning Agent first retrieves the user instruction and the initial screen from the device, then generates an initial plan using a Dual-ReAct formulation. This plan is transmitted to both the Execution Agent and the Observation Agent.

The Execution Agent and Observation Agent collaborate by alternately performing actions and evaluating outcomes based on step-wise expectations. At each step, the Observation Agent converts the screen into a compact semantic representation via the Pre-Understanding Module. When an unexpected scenario occurs, such as a permission popup not included in the original plan, the Observation Agent identifies the failure and transmits the corresponding textual representation to the cloud-based Memory Module. This information, together with the Reflection Module, assists the Planning Agent in performing replanning. The new plan is adapted to the current mobile environment and guides the two device-based agents to continue execution accordingly.

As shown in the example, most of the nine-step task is completed on-device. The cloud-based agent is triggered only twice, and aside from the initial planning, communication between device and cloud remains lightweight and textual, supporting low-bandwidth yet effective collaboration.

\subsection{Experimental Setup}
All device-side models are deployed on a local server with an NVIDIA RTX 3090 GPU (24G) to simulate mobile inference. We choose 2B- 4B scale models to ensure realistic feasibility for future mobile device deployment scenarios.

\subsection{System Overhead}
To characterize the computational and communication efficiency of EcoAgent, we report the 
per-task uplink data volume and the device-side forward FLOPs, following standard practice in 
device--cloud collaborative systems.

\begin{table}[h]
\centering
\begin{tabular}{lccc}
\toprule
Agent & AppAgent & M3A & EcoAgent (ShowUI) \\
\midrule
Data (kB) & 2098 & 5831 & \textbf{120} \\
\bottomrule
\end{tabular}
\caption{Per-task uplink data volume comparison of mobile agents.}
\label{tab:overhead}
\end{table}

In terms of computation, the device-side models require roughly \textbf{10--11 TFLOPs} per forward
pass (10.89 TFLOPs for Qwen2-VL-2B and 11.46 TFLOPs for ShowUI-2B), which aligns with the
capabilities of modern mobile NPUs.

As shown in Table~\ref{tab:overhead}, EcoAgent requires only \textbf{120 kB} of uplink data per 
task, which is \textbf{17.5$\times$} smaller than AppAgent and \textbf{48.6$\times$} smaller than 
M3A. This reduction primarily comes from performing step-wise verification and semantic 
summarization directly on the device, allowing the system to transmit only compact textual 
representations rather than raw screenshots.

Combined with the moderate device-side computational demand, these results demonstrate that 
EcoAgent achieves substantially lower system overhead compared with cloud-centric designs, 
making it highly suitable for deployment in future mobile automation scenarios.

\subsection{Failure Analysis}
To better understand the limitations of EcoAgent, we conduct a failure analysis based on 20 
randomly sampled unsuccessful EcoAgent (ShowUI) tasks. The failures are categorized into four 
types, as summarized in Table~\ref{tab:failure}.

\begin{table*}[h]
\centering
\begin{tabular}{lcccc}
\toprule
Failure Type & Planning & Visual Grounding & Max Steps & Verification \\
\midrule
Count & 7 & 10 & 2 & 1 \\
\bottomrule
\end{tabular}
\caption{Failure analysis of 20 randomly sampled unsuccessful EcoAgent (ShowUI) tasks.}
\label{tab:failure}
\end{table*}

Representative cases include:

\begin{itemize}
    \item \textbf{ExpenseAddSingle}: the Planning Agent misinterprets subgoals, leading to redundant or misaligned actions.
    \item \textbf{OsmAndMarker}: overlapping map elements introduce visual ambiguity for the 
Execution Agent, leading to incorrect action generation during marker selection.
\end{itemize}

As shown in Table~\ref{tab:failure}, approximately half of all failures (\textbf{10 out of 20}) 
stem from visual grounding errors. These issues arise primarily from the limited grounding 
capacity of MSLMs. Planning-related errors represent the second largest source, typically caused 
by subtle goal misinterpretation or insufficient contextual reasoning.

Overall, the dominance of grounding-related failures suggests that EcoAgent's performance can 
naturally improve as stronger device-based MSLMs become available. This trend highlights the 
\textbf{model-agnostic} nature of the framework: EcoAgent does not rely on any specific vision 
model architecture, and its device–cloud collaboration pipeline can directly benefit from future 
advances in lightweight mobile VLMs.

\subsection{Prompts}

In Table~\ref{plan}, ~\ref{replan}, ~\ref{preunderstand}, ~\ref{observation}, we present the prompts used by the cloud-based Planning Agent for both Dual-ReACT initial planning and Memory and Reflection empowered replanning, as well as those used by the device-based Observation Agent for pre-nderstanding and observation. The prompts for the Execution Agent follow the standard input format provided by the respective models and can be accessed via their Hugging Face model cards.
For all visual inputs, we use raw screen screenshots without any additional preprocessing.

\begin{table*}[h]
\centering
\begin{tabular}{p{0.95\linewidth}}
\toprule
\textbf{System:}\\
You are a mobile assistant that helps users complete tasks on Android devices.\\
\midrule
\textbf{User:}\\
Based on this Android screenshot, please help me with: \{instruction\}\\
\\
Please provide:\\
1. A description of what you see on the current screen\\
2. What needs to be done to achieve the goal\\
3. A step-by-step action plan and expectations in JSON format like this: \\
```JSON\\
\{\{\\
"Step1": \{\{"thought": "THOUGHT1", "step": "STEP1.", "expectation": "EXPECTATION1."\}\},\\
"Step2": \{\{"thought": "THOUGHT2", "step": "STEP2.", "expectation": "EXPECTATION2."\}\},\\
"Step3": \{\{"thought": "THOUGHT3", "step": "STEP3.", "expectation": "EXPECTATION3."\}\}\\
\}\}\\
```\\
\\
For each step, you can choose from the following actions:\\
1. TAP\\
2. SWIPE\\
3. INPUT\\
4. ENTER\\
5. ANSWER\\
6. OPEN\_APP\\
7. DELETE\\
\\
Make sure the step is brief and clear, you should only include the most important information in the step description. For example, if you need to tap a button, you should only outline the button's name but not its location on the screen.\\
\\
Your response MUST strictly adhere the following structure, and DO NOT add other contents:\\
\\
Description: $<$You need to describe the current screen and what you see on it.$>$\\
Thought: $<$You need to think about what needs to be done to achieve the goal.$>$\\
Plan: $<$You need to provide a plan in JSON format as described above.$>$ \\
\bottomrule
\end{tabular}
\caption{Dual-ReACT Initial Planning Prompt}
\label{plan}
\end{table*}

\begin{table*}[h]
\centering
\begin{tabular}{p{0.95\linewidth}}
\toprule
\textbf{System:}\\
You are a mobile assistant that helps users complete tasks on Android devices.\\
\midrule
\textbf{User:}\\
Given the original plan and current progress, please create a new plan for goal: \{instruction\}\\
\\
Original Plan:\\
\{original\_plan\}\\
\\
The screen has changed as follows:\\
\{descriptions\}\\
\\
The current screen description is: \{description\}\\
The reason why you failed at the last step is: \{summary\}\\
Please provide:\\
1. Why the previous action was not successful? What can you learn from this failure?\\
2. A step-by-step action plan started from this screen and expectations in JSON format like this: \\
```JSON\\
\{\{\\
"Step1": \{\{"thought": "THOUGHT1", "step": "STEP1.", "expectation": "EXPECTATION1."\}\},\\
"Step2": \{\{"thought": "THOUGHT2", "step": "STEP2.", "expectation": "EXPECTATION2."\}\},\\
"Step3": \{\{"thought": "THOUGHT3", "step": "STEP3.", "expectation": "EXPECTATION3."\}\}\\
\}\}\\
```\\
\\
For each step, you can choose from the following actions:\\
1. TAP\\
2. SWIPE\\
3. INPUT\\
4. ENTER\\
5. ANSWER\\
6. OPEN\_APP\\
7. DELETE\\
8. PRESS\_BACK\\
9. PRESS\_HOME\\
\\
Make sure the step is brief and clear, you should only include the most important information in the step description. For example, if you need to tap a button, you should only outline the button's name but not its location on the screen.\\
\\
Your response MUST strictly adhere the following structure, and DO NOT add other contents:\\
\\
Reflection: $<$You need to reflect on the last action and what you can learn from this failure.$>$\\
Plan: $<$You need to provide a new plan for the remaining steps in JSON format as described above.$>$ \\
\\
If you have reviewed the screen history CAREFULLY and confirmed that the task has been completed, please JUST respond with:\\
`No need to replan.'\\
\bottomrule
\end{tabular}
\caption{Memory and Reflection Empowered Replanning Prompt}
\label{replan}
\end{table*}

\begin{table*}[h]
\centering
\begin{tabular}{p{0.95\linewidth}}
\toprule
\textbf{System:}\\
You are an expert visual assistant trained to analyze smartphone screenshots.\\
\midrule
\textbf{User:}\\
- You are provided with a screenshot of a mobile phone.\\
- You are required to describe main contents and functionality of the current screen.\\
- You should focus on the main elements and their purposes, avoiding unnecessary details.\\
- Focus on brief, you should in 3-5 sentences.\\
\bottomrule
\end{tabular}
\caption{Pre-Understanding Module Prompt}
\label{preunderstand}
\end{table*}

\begin{table*}[h]
\centering
\begin{tabular}{p{0.95\linewidth}}
\toprule
\textbf{System:}\\
You are an intelligent and helpful visual assistant adept at analyzing mobile devices.\\
\midrule
\textbf{User:}\\
- You are required to observe the screenshot carefully.\\
user expectation: \{expectation\}\\
- You are required to judge whether the screenshot meets the user expectation based on your observation.\\
\hline
\end{tabular}
\caption{Observation Prompt}
\label{observation}
\end{table*}

\end{document}